\documentclass[dvipsnames,nonacm,format=sigconf,anonymous=false,review=false]{acmart}
\AtBeginDocument{%
  }

\usepackage[ruled,lined]{algorithm2e}
\usepackage{graphicx}
\usepackage{makecell}
\usepackage{cleveref}
\usepackage{float}
\usepackage{amsmath}
\begin{document}

\title{Detect and Act: Automated Dynamic Optimizer through Meta-Black-Box Optimization}

% \author{Anonymous Authors}

\author{Zijian Gao}
\email{gaozijian6812@gmail.com}
\orcid{0009-0008-6981-1836}
\affiliation{%
  \institution{South China University of Technology}
  \city{Guangzhou}
  \state{Guangdong}
  \country{China}
}

\author{Yuanting Zhong}
\email{ytalienzhong@gmail.com}
\orcid{0009-0004-5998-4822}
\affiliation{%
  \institution{South China University of Technology}
  \city{Guangzhou}
  \state{Guangdong}
  \country{China}
}

\author{Zeyuan Ma}
\email{scut.crazynicolas@gmail.com}
\orcid{0000-0001-6216-9379}
\affiliation{%
  \institution{South China University of Technology}
  \city{Guangzhou}
  \state{Guangdong}
  \country{China}
}

\author{Yue-Jiao Gong}
\email{gongyuejiao@gmail.com}
\orcid{0000-0002-5648-1160}
\affiliation{%
  \institution{South China University of Technology}
  \city{Guangzhou}
  \state{Guangdong}
  \country{China}
}

\author{Hongshu Guo}
\email{guohongshu369@gmail.com}
\orcid{0000-0001-8063-8984}
\affiliation{%
  \institution{South China University of Technology}
  \city{Guangzhou}
  \state{Guangdong}
  \country{China}
}
\authornote{Corresponding Author.}

% \affiliation{%

% \affiliation{%

% \affiliation{%

\renewcommand{\shortauthors}{Gao et al.}

\begin{abstract}
Dynamic Optimization Problems (DOPs) are challenging to address due to their complex nature, i.e., dynamic environment variation. Evolutionary Computation methods are generally advantaged in solving DOPs since they resemble dynamic biological evolution. However, existing evolutionary dynamic optimization methods rely heavily on human-crafted adaptive strategy to detect environment variation in DOPs, and then adapt the searching strategy accordingly. These hand-crafted strategies may perform ineffectively at out-of-box scenarios. In this paper, we propose a reinforcement learning-assisted approach to enable automated variation detection and self-adaption in evolutionary algorithms. This is achieved by borrowing the bi-level learning-to-optimize idea from recent Meta-Black-Box Optimization works. We use a deep Q-network as optimization dynamics detector and searching strategy adapter: It is fed as input with current-step optimization state and then dictates desired control parameters to underlying evolutionary algorithms for next-step optimization. The learning objective is to maximize the expected performance gain across a problem distribution. Once trained, our approach could generalize toward unseen DOPs with automated environment variation detection and self-adaption. To facilitate comprehensive validation, we further construct an easy-to-difficult DOPs testbed with diverse synthetic instances. Extensive benchmark results demonstrate flexible searching behavior and superior performance of our approach in solving DOPs, compared to state-of-the-art baselines.   
\end{abstract}

\begin{CCSXML}
<ccs2012>
   <concept>
       <concept_id>10010147.10010257.10010258.10010261.10010272</concept_id>
       <concept_desc>Computing methodologies~Sequential decision making</concept_desc>
       <concept_significance>500</concept_significance>
       </concept>
 </ccs2012>
\end{CCSXML}

\ccsdesc[500]{Computing methodologies~Sequential decision making}

\keywords{Dynamic Optimization, Particle Swarm Optimization, Parameter Control, Reinforcement Learning, Meta-Black-Box Optimization}


\maketitle

\section{Introduction}
Dynamic optimization problems~(DOPs) are optimization problems with a wide range of uncertainties \cite{1438403}. These uncertainties may affect objective functions, problem instances, and constraints \cite{NGUYEN20121}. The goal of DOPs is to minimize the difference between solutions found by the optimizer and optimal solutions at every optimization stage. DOPs are categorized into dynamic systems and dynamic environments. In systems (e.g., economic dispatch \cite{https://doi.org/10.1049/iet-gtd.2012.0726}, edge computing \cite{Rasool2024}), states evolve based on historical decisions. By contrast, environments (e.g., path planning \cite{XinOkayPlan}) are driven by external changes in objective landscapes or constraints. DOPs can be classified according to whether their search spaces are continuous or discrete,  targeted at single or multiple objectives, and whether they focus on tracking optimal solutions or finding solutions that are robust in the case of future environmental changes \cite{Yazdani2025}. 

Evolutionary algorithms~(EAs) and swarm intelligence~(SI) are inspired by natural evolution, which is a continuous adaptation process. Therefore, they are suitable candidates for solving DOPs \cite{1438403}. However, uncertainties in DOPs pose two primary challenges for standard EAs and SI: the invalidation of accumulated search information and the diversity loss due to species convergence. To address these, specific memory mechanisms have been introduced \cite{CAO2018463, LUO2016130}. While convergence is desired in static landscapes, it restricts the algorithm's ability to react to environmental changes. Thus, multi-population \cite{1665033, LI201595} and diversity maintenance mechanisms \cite{Blackwell2002} are introduced to sustain exploratory ability. These improvements have led to the development of Dynamic Optimization Algorithms~(DOAs). According to the classification in \cite{8657680}, the core components of DOAs include change detection, diversity control, and population management. Among these, change detection is a prerequisite for any adaptive response, generally implemented via reevaluation-based \cite{785502} or fitness monitoring-based methods \cite{Bb2016}.

Although recent works have obtained some promising results, two bottlenecks remain: 1) These methods, which design specialized components tailored to specific DOPs, are constrained by the no-free-lunch theorem \cite{585893}. What works for one problem type may fail for another, resulting in limited generalization. 2) These methods depend on hand-crafted mechanisms to detect environmental changes and execute responses. Such a detect-then-act pipeline requires complex manual design and lacks the efficiency of an end-to-end framework.
As an emerging field, Meta-Black-Box-Optimization (MetaBBO) leverages bi-level learning-assisted frameworks to automate algorithm design, e.g. algorithm configuration where the hyper-parameters and/or operators of the low-level algorithm are adjusted by the meta-level policy to adapt for the given problem \cite{10993463}, significantly reducing manual expertise costs.

While existing MetaBBO frameworks are predominantly evaluated on stationary landscapes, their inherent feedback mechanism makes them well-suited to the dynamic nature of DOPs. Motivated by the success of bi-level learning architectures and the concept of learning-to-optimize in recent MetaBBO studies, this paper presents a reinforcement learning-assisted system that streamlines the detect-then-act pipeline and automates it as a Markov Decision Process~(MDP). By doing so, our system is end-to-end, enabling the current optimization state to be mapped directly to strategy adjustment in a single step, thereby eliminating the need for hand-crafted change detection mechanisms and response heuristics, achieving superior performance across various DOPs.

We now summarize the contributions in this paper:
\begin{itemize}
\item[$\bullet$] Introduction of Meta-DO, an end-to-end reinforcement learning framework that replaces traditional hand-crafted detect-then-act pipelines with direct state-to-strategy mapping.
\item[$\bullet$] A comprehensive state representation leveraging an elite archive to perceive landscape dynamics and environmental drift without the need for explicit change detectors.
\item[$\bullet$] A log-scaled reward scheme and joint control of $w, c_1 \text{ and } c_2$ hyper-parameters, ensuring stable learning and flexible adaptation in non-stationary environments.
\item[$\bullet$] Extensive validation on 32 benchmark instances and a real-world USV navigation task, demonstrating superior performance and cross-domain generalization capability.
\end{itemize}
\section{Related Works}
\subsection{Dynamic Optimization}
Formally, a single-objective DOP can be defined as: $f^{(t+1)}=\mathcal{V}\circ f^{(t)}$, where $f$ is the objective function, $t\in\{1,\dots,T_{\text{max}}\}$ is the time index and $\mathcal{V}$ is the dynamic transition introducing temporal uncertainties. To maintain performance, traditional algorithms require specialized components for memory \cite{CAO2018463}, multi-population management \cite{LI201595}, and diversity maintenance \cite{Blackwell2002}. A critical prerequisite is change detection, which is generally categorized into reevaluation-based and fitness monitoring-based methods \cite{9356715}. Reevaluation\-based methods select detectors from a variety of sources, including  archived/fixed points \cite{785502, 6076335}, randomly chosen points \cite{1049555, DU20083096, 10.1007/978-3-319-77538-8_56} and best found points from subpopulations \cite{10.1007/978-3-540-24653-4_50}. Fitness monitoring based methods track evaluated fitness values, including the fitness difference between best individuals \cite{1004492}, average fitness values \cite{Bb2016} and fitness value fluctuations \cite{10.1007/978-3-030-26369-0_33}. 

Additionally, the response mechanisms following the detection of a change also rely on hand-crafted reactive logic, such as reinitialization \cite{10.1145/1143997.1144187}, optimal particles calibration and diversity maintenance \cite{LIU2020105711, LI2024121254} and modify the shift severity within the introducing diversity mechanism \cite{8657680}. Since reinitialization may waste valuable historical information, subsequent research has begun to shift toward memory-based or prediction based response mechanisms, attempting to preserve useful information by archiving \cite{10.1007/978-3-540-71805-5_70} (randomly placing particles near archive solutions/replacing the worst particles with archive solutions) or predicting optimum trends and transferring historical solutions to new promising solutions \cite{8767010, 8536381, ZHANG2025102011}. However, these hand-crafted detect-then-act pipelines depend heavily on expert knowledge and intensive parameter tuning, leading to limited robustness in complex landscapes. This necessitates a shift toward automated, end-to-end frameworks that can directly perceive environmental dynamics and coordinate adaptation strategies without manual intervention.

\subsection{Meta-Black-Box Optimization}
Existing MetaBBO research focuses on automating BBO algorithms design through various learning paradigms:
1) Reinforcement learning~(MetaBBO-RL): training a meta-policy to decide sequentially for algorithmic configurations \cite{ma2024auto, 9359652, 10496708, ma2025metablackboxoptimization, ma2025surrogate, guo2025reinforcement, guo2025designx, guo2025advancing};
2) Auto-regre\-ssive supervised learning~(MetaBBO-SL): Learning to predict optimization steps or parameters by imitating expert trajectories \cite{pmlr-v70-chen17e, ma2024llamocoinstructiontuninglarge, lange2024evolutiontransformerincontextevolutionary, li2025b2opt};
3) Neuroevolution~(MetaBBO-NE): Using evolutionary strategies at the meta-level to evolve the architecture or weights of the optimizer's neural network \cite{10.1145/3583133.3595822, 10.1145/3583131.3590496, li2025meta};
4) Large Language Models (LLMs)-based in-context learning~(MetaBBO-ICL): Leveraging LLMs to suggest optimization moves or code components based on historical search data in a zero-shot or few-shot manner \cite{guo2025evopromptconnectingllmsevolutionary, liu2024evolutionheuristicsefficientautomatic, mo2025autosgnn, huang2025evaluation, zhang2025systematic,han2025enhancing,li2024pretrained,ahmaditeshnizi2024optimusscalableoptimizationmodeling}.

Specifically, MetaBBO-RL models algorithm configuration as an MDP within a bi-level framework. The meta-level agent interacts with the low-level optimizer to autonomously determine the most suitable search behaviors. At step $t$, the agent observes state $s_t$ (landscape \cite{wright1932roles, 10.1145/2001576.2001690, ma2024neural} and population features) and selects action $a_t$ (e.g. hyper-parameter tuning \cite{9359652, Wu01022022} or operator selection \cite{TAN2021107678}) via policy $\pi_\theta$ to adjust the optimizer. The resulting performance gain yields reward $r_t$ and transition to $s_{t+1}$. These trajectories are used to train the agent based on reinforcement learning rules, with the aim of maximizing the expected cumulative reward and thus yielding a robust policy for automated algorithm design \cite{10.1145/3638529.3653995}.

Despite its impressive performance in invariant environments, a major limitation of current MetaBBO is its reliance on the assumption of stationarity. Most existing MetaBBO frameworks are designed to generalize across different static problem instances, rather than adapting to temporal dynamics within a single optimization process. This research gap motivates the development of dynamic-aware MetaBBO approaches that can bridge the divide between automated algorithm design and volatile real-world DOPs.

\section{Methodology}
\subsection {Dynamic Optimization Problem}
A single-objective DOP can be defined as $f^{(t+1)}=\mathcal{V}\circ f^{(t)}$, where $\mathcal{V}$ represents the dynamic transition. For example, in the Generalized Moving Peaks Benchmark~(GMPB) \cite{DBLP:journals/corr/abs-2106-06174}, the transition operator $\mathcal{V}$ is represented as a set of stochastic update rules applied to the $N_c$ components that constitute the fitness landscape. Specifically, the landscape at time $t$ is determined by a vector of time-varying parameters $\mathcal{P}^{(t)} = \{o_l^{(t)}, \eta_l^{(t)}, \omega_l^{(t)}, \varphi_l^{(t)}, \dots\}_{l=1}^{N_c}$, representing the centers, heights, widths, and rotation angles of the peaks.
In our implementation, $\mathcal{V}$ denotes the environmental changes, which include introducing linear additive noise to the fitness values, shifting the fitness function, and their hybrid combinations.

The objective of DOPs is to minimize the offline error ($E_{\text{off}}$), which measures the algorithm's ability to track the moving optimum over the entire optimization process. For the $\tau$-th function evaluation, let $x^{*,\tau}$ represent the position of the global optimum and $x^{\text{best},\tau}$ denote the best solution found by the algorithm up to that point. The optimization goal is formulated as:
\begin{equation}
\text{minimize:} E_{\text{off}} = \frac{1}{\mathit{FE}_{\text{max}}} \sum_{\tau=1}^{\mathit{FE}_{\text{max}}} \left(f^{(\tau)}(x^{\text{best},\tau}) - f^{(\tau)}(x^{*,\tau})\right)
\end{equation}
where $\mathit{FE}_{\text{max}}$ is the maximum budget of function evaluations~(FEs) and $f^{(\tau)}$ is the fitness function at the $\tau$-th function evaluation. This metric is particularly significant as it encapsulates not only the algorithm's convergence speed within a single environment but also its adaptability and recovery rate following environmental changes governed by the transition operator $\mathcal{V}$.

\subsection {Low-level Optimizer: NBNC-PSO}
To maintain multiple niches and track local optima in shifting landscapes, we use NBNC-PSO \cite{9295420} as the low-level optimizer. NBNC is a clustering-based niching method that groups species together by linking individuals to their nearest-better neighbours within a defined proximity. Distinct from traditional nearest-better methods, NBNC characterizes the spatial relationship of individuals by calculating distances to their nearest neighbors rather than solely relying on their nearest-better counterparts to determine cluster boundaries. To refine these initial clusters, a merging mechanism based on a defined dominance relationship between species is introduced. Under this scheme, a dominated raw species is systematically merged into its dominator, eventually forming the final species in the population. Integrating this clustering logic into the Particle Swarm Optimization~(PSO) framework results in an enhanced NBNC-PSO algorithm with improved exploration capabilities. This enables the algorithm to maintain multiple niches and effectively track local optima in complex landscapes.

In order to provide the meta-level policy with high-quality historical references, we have supplemented the standard NBNC-PSO with an elite archive. This archive stores the best-found individuals from the past five generations. Specifically, the archive $Ar$ serves as a key mechanism for environmental detection. Specifically, an environmental scaling factor $\mathit{ratio}$ is introduced to characterize landscape dynamics without explicit change detection, as defined in \Cref{eq:ratio}. This factor is taken into the state representation to represent the magnitude of environmental drift. Furthermore, $\mathit{ratio}$ facilitates reward normalization by providing a reference for performance improvement as detailed in \Cref{sec:reward}. 
\subsection{Meta-level RL Framework}
\begin{figure}[t]
    \centering
    \includegraphics[width=\columnwidth]{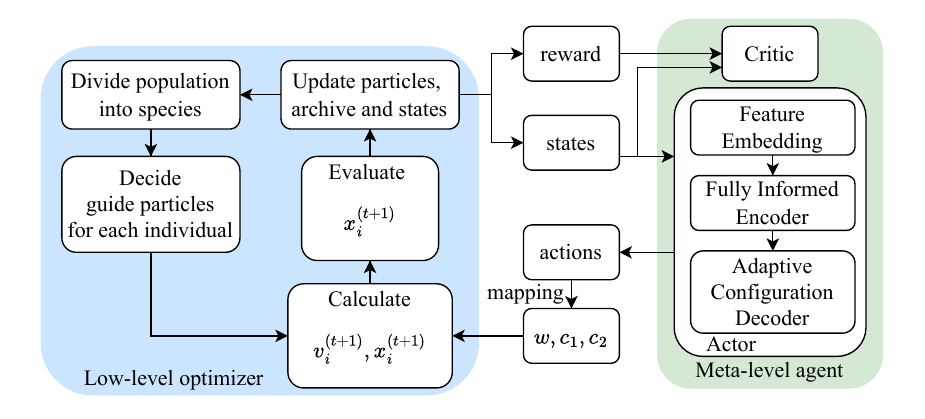}
    \vspace{-3mm}
    \caption{Flowchart of Meta-DO}
    \vspace{-3mm}
    \label{Fig.flowchart}
\end{figure}
The proposed approach adopts a bi-level learning structure, where the meta-level reinforcement learning~(RL) agent autonomously configures the low-level NBNC-PSO optimizer using the architecture introduced in \cite{ma2024auto}. By adopting a dynamic perspective on the evolutionary process, the RL agent learns a generalized policy that enables it to dynamically adjust its searching strategies in response to environmental changes caused by $\mathcal{V}$.
\subsubsection{MDP Formulation}
Given a population $P$ with $N$ individuals, an algorithm $\Lambda$ and a problem set $\mathcal{D}$, the dynamic tuning process is formulated as an MDP: 
\begin{equation}
\mathcal{M}:=\langle \mathcal{S}=\{s_i\}^N_{i=1}, \mathcal{A}=\{a_i\}^N_{i=1}, \mathcal{T}, \mathcal{R} \rangle
\end{equation}
where state $\mathcal{S}$ and action $\mathcal{A}$ consider all individuals in the $P$. Each $a_i\in \mathbb{R}^M$ represents the configuration of $M$ hyper-parameters for the $i$-th individual. In this study, the agent determines the optimal values for the inertia weight ($w$), the cognitive coefficient ($c_1$), and the social coefficient ($c_2$) as shown in \Cref{Fig.flowchart} (bottom-center).

The transition function $\mathcal{T}: \mathcal{A} \times \Lambda \times P \rightarrow P$ represents the evolution of the population through the algorithm $\Lambda$ under the influence of the dynamic transition operator $\mathcal{V}$. The reward function $\mathcal{R}: \mathcal{S} \times \mathcal{A} \times \mathcal{D}\rightarrow \mathbb{R}$ measures the improvement in one optimization step brought by dynamic hyper-parameter settings.
\subsubsection{State} \label{sec:state}
% Inspired by the design principles in GLEET \cite{ma2024auto}
The state $\mathcal{S} \in \mathbb{R}^{N \times \mathcal{K}}$ is constructed to provide the RL agent with a comprehensive perception of the population status and environmental dynamics. In this study, the state vector for each individual $i$ is defined as a 10-dimensional feature vector ($\mathcal{K}=10$), which can be categorized into five functional groups:
\begin{itemize}
\item[1)] Environmental Variation Perception ($\mathit{fea}_1$)

This feature quantifies the landscape shift intensity by calculating the log-scaled ratio of the reevaluated fitness of historical optima. It serves as a change detector for non-stationary environments:
\begin{equation} \label{eq:ratio}
\mathit{ratio} = \text{mean}\left(\frac{f(Ar^{(t)}) + \epsilon}{f(Ar^{(t-1)}) + \epsilon}\right)
\end{equation}
\begin{equation}
\mathit{fea}_1 = \text{clip}\left(\frac{\log_{10}\mathit{ratio}}{8}, -1, 1\right)
\end{equation}
where $f(Ar^{(t)})$ and $f(Ar^{(t-1)})$ represent the fitness values of the historical best positions stored in the archive at time steps $t$ and $t-1$, respectively. The parameter $\epsilon=10^{-8}$ is a constant added to ensure numerical stability for the division and the subsequent logarithmic scaling.
\item[2)] Global and Local Fitness Distribution ($\mathit{fea}_2, \mathit{fea}_3$)

\textbf{\boldmath$\mathit{fea}_2$ (Global Normalized Fitness)}: This feature characterizes the relative performance of an individual within the entire population. It is formally defined as:
\begin{equation}
fea_{2,i} = \frac{f(x_i) - \text{mean}(\{f(x_i)\}_{i=1}^N)}{\text{std}(\{f(x_i)\}_{i=1}^N) + \epsilon}
\end{equation}
where $f(x_i)$ denotes the fitness of individual $i$.

\textbf{\boldmath$\mathit{fea}_3$ (Local Normalized Fitness)}: To capture the local convergence status within different niches, we introduce a species-level standardized feature:
\begin{equation}
\mathit{fea}_{3,i} = \frac{f(x_i) - \text{mean}(\{f(x_i)\}_{x_i\in S_j})}{\text{std}(\{f(x_i)\}_{x_i\in S_j}) + \epsilon}, \quad \text{for } x_i \in S_j
\end{equation}
where $S_j$ denotes the $j$-th species containing individual $i$, as identified by the NBNC process. This local feature enables the agent to evaluate an individual's proficiency relative to its neighbors, facilitating precise local search control.
\item[3)] Search Progress and Stagnation ($\mathit{fea}_4, \mathit{fea}_5, \mathit{fea}_6$)

\textbf{\boldmath$\mathit{fea}_4$ (Search Horizon)}: The ratio of remaining FEs: \begin{equation} \mathit{fea}_4 = \frac{\mathit{FE}_{\text{max}} - \mathit{FE}}{\mathit{FE}_{\text{max}}} \end{equation}
where $\mathit{FE}$ is the currently consumed number of FEs.

\textbf{\boldmath$\mathit{fea}_5$ and \boldmath$\mathit{fea}_6$ (Stagnation Status)}: These features represent the stagnation status regarding the number of generations $z(\cdot)$ for which the algorithm failed to obtain a better personal best ($\mathit{pBest}_i$) or global best ($\mathit{gBest}$), respectively, normalized by the maximum generation limit $T_{\text{max}}=\frac{FE_{\text{max}}}{N}$.
\begin{equation}
\mathit{fea}_5 = \frac{z(\mathit{pBest}_i)}{T_{\text{max}}}, \quad \mathit{fea}_6 = \frac{z(\mathit{gBest})}{T_{\text{max}}}
\end{equation}
\item[4)] Spatial Topology ($\mathit{fea}_7, \mathit{fea}_8, \mathit{fea}_9$)

\textbf{\boldmath$\mathit{fea}_7, \mathit{fea}_8$ (Distance to Global and Local Best)}: \begin{equation}
\mathit{fea}_7 = \frac{||x_i - \mathit{gBest}||}{\mathit{diameter}}, \quad \mathit{fea}_8 = \frac{||x_i - \mathit{sBest}_j||}{\mathit{diameter}}
\end{equation}
where $\mathit{diameter} = \sqrt{\sum^D_{d=1} (U_d - L_d)^2}$ is the maximum possible distance within the search space. $L$ and $U$ denote the lower and upper bound vectors, respectively, and $D$ is the dimensionality of the search space. $\mathit{sBest}_j$ is the best individual within the $j$-th species.

\textbf{\boldmath$\mathit{fea}_9$ (Distance to Personal Best)}: 
\begin{equation} \mathit{fea}_9 = \frac{||x_i - \mathit{pBest}_i||}{\mathit{diameter}} \end{equation}
\item[5)] Directional Correlation ($\mathit{fea}_{10}$)

This feature uses cosine similarity to measure the alignment between the vectors pointing toward the personal best and the global best: \begin{equation} \mathit{fea}_{10} = \cos{\left(\angle(\mathit{gBest}-x_i,\mathit{pBest}_i-x_i)\right)} \end{equation} It indicates whether the cognitive and social components are providing conflicting or consistent search directions.
\end{itemize}
\subsubsection{Action} \label{sec:action}
Instead of discretizing the action space which may damage the action structure or trigger the curse of dimensionality \cite{lillicrap2019continuouscontroldeepreinforcement}, our framework adopts a continuous action space that jointly determines the hyper-parameters $(a^{(t)}_1,a^{(t)}_2,\dots,a^{(t)}_N)$ for all $N$ individuals, where $a^{(t)}_i$ denotes $M$ hyper-parameters for individual $i$ at time step $t$. 
Formally, the action probability $\text{Pr}(a)$ is modeled as a product of normal distributions as follows:
\begin{equation}
\text{Pr}(a)=\prod^N_{i=1}\prod^M_{m=1}p(a^m_i),\quad a^m_i\sim \mathcal{N}(\mu^m_i,\sigma^m_i)
\end{equation}
where $\mu^m_i$ and $\sigma^m_i$ are controlled by the RL agent. In our implementation, $M=3$ and the policy network outputs $N\times 3$ pairs of $(\mu^m_i,\sigma^m_i)$. Specifically, $\mu^m_i \in [0, 1]$ and $\sigma^m_i \in [10^{-3}, 0.7]$ via $tanh$ activation and linear scaling. Actions $a^m_i$ are sampled from the normal distribution $\mathcal{N}(\mu^m_i, \sigma^m_i)$ to maintain exploration in a continuous space. To ensure validity, the sampled actions $a^{(t)}_i \in [0, 1]^M$ are mapped to the actual hyper-parameter vector $h_i^{(t)} = [w, c_1, c_2]^\top$ via a linear transformation: $h_{i}^{(t)}=\text{diag}(U-L)\cdot a_{i}^{(t)}+L$.
\subsubsection{Reward} \label{sec:reward}
The reward $r_t$ is designed to provide a stable and meaningful learning signal despite the non-stationary nature of DOPs. Essentially, the reward encourages the agent to achieve a fitness improvement that exceeds the natural drift of the landscape. The calculation involves three steps:
\begin{itemize}
\item[1)] Environmental Alignment:
In DOPs, a change in fitness value may be caused by a landscape shift rather than the optimizer's progress. To measure the optimizer's actual performance, we use the scaling factor $\mathit{ratio}$ (from \Cref{eq:ratio}) to estimate the current environment's prior best: 
\begin{equation}
f_{\text{base}}^{(t)} = \mathit{ratio} \times f(\mathit{gBest}^{(t-1)})
\end{equation}
This $f_{\text{base}}^{(t)}$ serves as an adjusted baseline, representing the expected performance if the algorithm made no progress under the new environment.
\item[2)] Log-Fitness Improvement:
To ensure the agent is equally sensitive to improvements across magnitudes (e.g., from $10^6$ to $10^{-8}$), we calculate the improvement $\Delta$ in log-space:
\begin{equation}
\Delta = \log_{10}\left(\text{max}(|f_{\text{base}}^{(t)}|,\epsilon)\right) - \log_{10}\left(\text{max}(|f(\mathit{gBest}^{(t)})|,\epsilon)\right)
\end{equation}
A positive $\Delta$ indicates that the optimizer has outperformed the environmental drift.
\item[3)] Normalization for Training Stability:
Finally, to ensure the reward signal remains within a consistent numerical range whatever the scale of the problem, we normalize $\Delta$ by the current maximum potential log-improvement:
\begin{equation}
r_t=\frac{\text{max}(\Delta, 0)}{\log_{10}\left(\text{max}(f_{\text{base}}^{(t)},\epsilon)\right)-\log_{10}\epsilon+\epsilon}
\end{equation}
This normalization maps the reward to a relative scale, preventing destabilizingly large policy updates and facilitating smoother convergence in PPO training.
\end{itemize}
\subsubsection{Network design}
\begin{figure}[t]
    \centering
    \includegraphics[width=0.9\columnwidth]{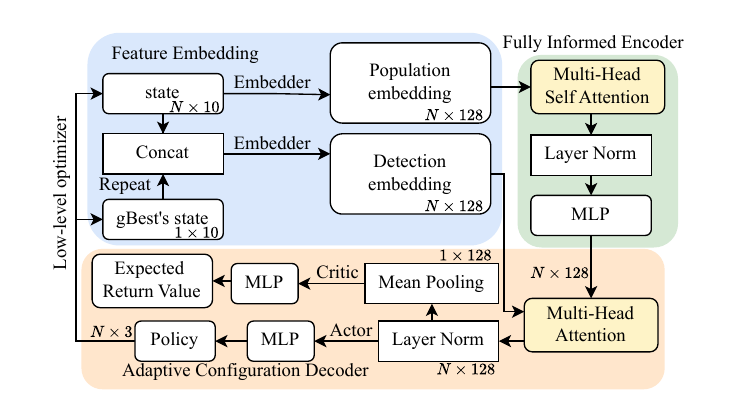}
    \vspace{-3mm}
    \caption{Illustration of our network}
    \vspace{-5mm}
    \label{Fig.network}
\end{figure}
To process population-level features and derive optimal hyper-parameter configurations, our framework adopts the Transformer-styled network architecture \cite{ma2024auto}. As shown in \Cref{Fig.network}, the network follows an actor-critic paradigm, where the actor $\pi_{\theta}$ consists of three modules: a feature embedding module, a fully informed encoder, and an adaptive configuration decoder.
\begin{itemize}
\item[$\bullet$] Feature Embedding Module (\Cref{Fig.network}, top-left): The 10-dimen\-sional state features in \Cref{sec:state} are linearly projected into two groups of 128-dimensional embeddings: population embeddings~(PEs) and detection embeddings~(DEs) summarizing detection status.
\item[$\bullet$] Fully Informed Encoder (\Cref{Fig.network}, top-right): To facilitate information exchange among individuals, we employ a single-layer Transformer encoder. Consistent with \cite{ma2024auto}, the Queries (Q), Keys~(K), and Values~(V) within each layer are all derived from the PEs, enabling the swarm to adaptively adjust its internal topology. This architecture ensures permutation invariance, allowing Meta-DO to generalize across varying population sizes. The final output
is the fully informed population embeddings~(FIPEs).
\item[$\bullet$] Adaptive Configuration Decoder (\Cref{Fig.network}, bottom): This module derives the joint distribution of hyper-parameters through a Multi-Head Attention mechanism. Specifically, the DEs serve as the $Q$ to obtain individualized information from FIPEs (acting as $K$ and $V$). The output distributions $(\mu, \sigma)$ are then used to sample continuous actions as formulated in \Cref{sec:action}.
\item[$\bullet$] Critic Network: The critic network $\nu_{\psi}$ shares the feature representations from the actor but utilizes its own MLP layers. It performs mean pooling over the population's logits to estimate the state value, thereby assisting the training of the actor via PPO \cite{schulman2017proximalpolicyoptimizationalgorithms}.
\end{itemize}

\subsection{Implementation and Training}
\begin{algorithm}[htbp]
\caption{Meta-Training of Meta-DO Framework} \label{alg:train}
\KwIn{Policy $\pi_{\theta}$, Critic $\nu_{\psi}$, Training problem set $\mathcal{D}_\text{train}$}
\KwOut{Trained Policy $\pi_{\theta}$, Critic $\nu_{\psi}$}
\For{epoch = 1 \KwTo Epoch}{
    \For{$f \in \mathcal{D}_\text{train}$}{
        Initialize population $P$ and archive $Ar \gets \emptyset$\;
        Extract initial state $s_1$ from $P$\;
        \For{$t=1$ \KwTo $T_{\text{max}}$}{
            Sample $a_t \sim \pi_{\theta}(s_t)$ and map to $(w, c_1, c_2)$\;
            Update $P$ via one step of NBNC-PSO \cite{9295420}\;
            Update $gBest$ and $Ar$\;
            Replace the worst individual in each species with the best solution in $Ar$\;
            Re-evaluate $Ar$ and compute reward $r_t$\;
            Extract next state $s_{t+1}$ from $P$\;
            Store transition $(s_t,a_t,r_t,s_{t+1})$\;
            \If{$t\pmod n==0$}{
                \For{$k = 1$ \KwTo $\kappa$}{
                    Update $\pi_{\theta}, \nu_{\psi}$ by PPO method\;
                }
            }
        }
    }
}
\end{algorithm}
As illustrated in \Cref{alg:train}, the framework follows a bi-level loop. During the Inference phase, the algorithm first partitions the population into species using the NBNC operator to perceive local landscape features. The RL agent then extracts a 10-dimensional state vector for each individual and dictates the continuous hyper-parameters $(w, c_1, c_2)$ to the low-level NBNC-PSO. In our implementation, the policy is updated every $n=10$ steps, and for each update, the PPO optimization is performed for $\kappa=3$ epochs to ensure stable convergence.

To ensure robust generalization, the meta-policy is trained on a distribution of 64 non-stationary problem instances, generated using the mechanisms (weighted composition and stochastic noise) described in \Cref{sec:benchmark-generation}. During training, we follow the $T$-step PPO paradigm \cite{schulman2017proximalpolicyoptimizationalgorithms} to collect trajectories and update the actor-critic networks on-the-fly. This integrated approach allows the agent to learn a generalized strategy for diverse environmental transitions.
\section{Experiments}
\subsection{Experimental Setup}
To evaluate Meta-DO, we select eight SOTA algorithms specifically designed for DOPs, covering diverse strategies:
1) DE-based method: DynDE \cite{1555047} for diversity maintenance through multi-type individuals.
2) Multi-population frameworks: mDE and mCMAES \cite{8657680}, 
which employ sub-population pools to track multiple peaks simultaneously.
3) Adaptive \& Speciation-based PSO: ACFPSO \cite{9284465} for adaptive role management; PSPSO \cite{signorelli2025perturbationspeciationbasedalgorithmdynamic} for handling dynamics without explicit detection; and NBNC-PSO \cite{9295420}, a nearest-better clustering niching PSO that also serves as the optimization core of our framework.
4) Clustering-based PSO: APCPSO \cite{LIU2020105711} and DPCPSO \cite{LI2024121254}, which utilize Affinity Propagation and Density Peak Clustering for autonomous population division.

All baselines are reimplemented in Python within a unified framework to ensure a fair comparison, strictly following the logic and parameters settings specified in \cite{signorelli2025perturbationspeciationbasedalgorithmdynamic, 10.1145/3785134}. All algorithms are evaluated across identical test instances, with mean results reported over 10 independent runs to ensure statistical significance. More details can be found in Appendix \ref{appx:A}. %A.
\subsection{Benchmark generation} \label{sec:benchmark-generation}
Our experiments are conducted using a customized dynamic optimization suite, implemented via the \texttt{Dynamic}\texttt{\_Dataset} class. Unlike traditional static benchmarks, this framework constructs complex non-stationary landscapes through three key mechanisms:
1) \textbf{Landscape Switching}: The environment undergoes discrete transitions between different base problems (e.g., Sphere and Ackley). At any time $t$, the fitness is determined by a single active sub-problem, simulating abrupt structural shifts in the landscape topology.
2) \textbf{Temporal Scheduling}: Our mechanism controls the timing and frequency of transitions. By triggering these shifts periodically, it forces the algorithm to perceive and respond to temporal dynamics.
3) \textbf{Stochastic Noise}: To simulate real-world measurement uncertainties, a time-varying Gaussian noise is injected into the fitness evaluation. The noise intensity increases dynamically as the optimization proceeds.

Using the \texttt{Dynamic}\texttt{\_Dataset} generator, we produced 96 unique problem instances, partitioned into a training set $\mathcal{D}_\text{train}$ (64 instances) and a disjoint testing set (32 instances). All algorithms are evaluated on the 32 test instances ($f_{1}\text{-}f_{32}$), which are categorized by transition complexity: 1) Pure Noise ($f_{1}\text{-}f_{14}$) for basic change detection capability under stochastic interference; 2) Landscape Switching ($f_{15}\text{-}f_{24}$) for adaptation to structural landscape transitions; 3) Hybrid Transitions ($f_{25}\text{-}f_{32}$) for evaluating long-term robustness under coupled dynamic-topological complexities.
\subsection{Comparison Analysis}
To ensure a fair comparison across heterogeneous test instances with varying scales and complexities, all numerical results are presented as Normalized Performance Ratios. Specifically, for each instance, the raw offline error $E_{\text{off}}$ is normalized by the mean performance of 100 randomly sampled points $E_{\text{rand}}$, formulated as: $RP = \frac{E_{\text{off}}}{E_{\text{rand}}}$.
A lower $RP$ value indicates a higher degree of optimization efficiency relative to a random search baseline.

\begin{table*}[htbp]
\centering
\caption{Comparative Results on Benchmark}
\label{tab:full_res_on_benchmark}
\resizebox{\textwidth}{!}{
\begin{tabular}{c | cc | cc | cc | cc | cc | cc | cc | cc | cc } \hline
\text{Algorithm} & \multicolumn{2}{c|}{\textbf{Meta-DO}} & \multicolumn{2}{c|}{NBNC-PSO} & \multicolumn{2}{c|}{PSPSO} & \multicolumn{2}{c|}{ACFPSO} & \multicolumn{2}{c|}{mCMAES} & \multicolumn{2}{c|}{mDE} & \multicolumn{2}{c|}{APCPSO} & \multicolumn{2}{c|}{DPCPSO} & \multicolumn{2}{c}{DynDE} \\ \hline
Metrics & \makecell{Mean \\ (Std)} & Rank & \makecell{Mean \\ (Std)} & Rank & \makecell{Mean \\ (Std)} & Rank & \makecell{Mean \\ (Std)} & Rank & \makecell{Mean \\ (Std)} & Rank & \makecell{Mean \\ (Std)} & Rank & \makecell{Mean \\ (Std)} & Rank & \makecell{Mean \\ (Std)} & Rank & \makecell{Mean \\ (Std)} & Rank \\ \hline
$f_{1}$ & \textbf{\makecell{1.342e-02 \\ ($\pm$1.357e-03)}} & 1 & \makecell{3.399e-02 \\ ($\pm$2.266e-03)} & 2 & \makecell{4.758e-01 \\ ($\pm$7.323e-02)} & 5 & \makecell{2.319e-01 \\ ($\pm$2.484e-01)} & 3 & \makecell{1.038e+00 \\ ($\pm$2.285e-01)} & 9 & \makecell{9.246e-01 \\ ($\pm$1.556e-01)} & 7 & \makecell{7.015e-01 \\ ($\pm$3.233e-03)} & 6 & \makecell{4.633e-01 \\ ($\pm$8.542e-03)} & 4 & \makecell{9.800e-01 \\ ($\pm$2.445e-02)} & 8 \\
$f_{2}$ & \textbf{\makecell{1.210e-02 \\ ($\pm$2.032e-03)}} & 1 & \makecell{3.209e-02 \\ ($\pm$5.552e-03)} & 2 & \makecell{3.379e-01 \\ ($\pm$5.910e-02)} & 4 & \makecell{5.071e-02 \\ ($\pm$3.122e-02)} & 3 & \makecell{7.576e-01 \\ ($\pm$1.739e-01)} & 7 & \makecell{7.6048e-01 \\ ($\pm$3.456e-01)} & 8 & \makecell{4.656e-01 \\ ($\pm$4.379e-03)} & 6 & \makecell{4.583e-01 \\ ($\pm$4.440e-03)} & 5 & \makecell{9.662e-01 \\ ($\pm$8.516e-02)} & 9 \\
$f_{3}$ & \textbf{\makecell{9.885e-03 \\ ($\pm$1.035e-03)}} & 1 & \makecell{2.946e-02 \\ ($\pm$8.015e-03)} & 2 & \makecell{3.768e-01 \\ ($\pm$6.273e-02)} & 4 & \makecell{1.331e-01 \\ ($\pm$8.674e-02)} & 3 & \makecell{1.119e+00 \\ ($\pm$3.885e-01)} & 9 & \makecell{8.182e-01 \\ ($\pm$1.572e-01)} & 7 & \makecell{5.130e-01 \\ ($\pm$6.099e-03)} & 6 & \makecell{4.629e-01 \\ ($\pm$7.830e-03)} & 5 & \makecell{1.006e+00 \\ ($\pm$7.060e-02)} & 8 \\
$f_{4}$ & \textbf{\makecell{9.671e-02 \\ ($\pm$2.956e-02)}} & 1 & \makecell{3.543e-01 \\ ($\pm$2.296e-02)} & 2 & \makecell{8.074e-01 \\ ($\pm$5.253e-02)} & 5 & \makecell{6.906e-01 \\ ($\pm$1.434e-01)} & 3 & \makecell{9.768e-01 \\ ($\pm$6.323e-02)} & 8 & \makecell{9.570e-01 \\ ($\pm$5.581e-02)} & 7 & \makecell{8.748e-01 \\ ($\pm$2.858e-03)} & 6 & \makecell{6.912e-01 \\ ($\pm$3.832e-02)} & 4 & \makecell{9.911e-01 \\ ($\pm$7.757e-03)} & 9 \\
$f_{5}$ & \textbf{\makecell{4.413e-01 \\ ($\pm$3.447e-02)}} & 1 & \makecell{6.238e-01 \\ ($\pm$2.976e-02)} & 3 & \makecell{8.846e-01 \\ ($\pm$3.736e-02)} & 6 & \makecell{5.990e-01 \\ ($\pm$1.100e-01)} & 2 & \makecell{1.013e+00 \\ ($\pm$2.652e-02)} & 9 & \makecell{1.001e+00 \\ ($\pm$4.749e-02)} & 7 & \makecell{7.977e-01 \\ ($\pm$7.847e-03)} & 5 & \makecell{7.401e-01 \\ ($\pm$2.870e-02)} & 4 & \makecell{1.010e+00 \\ ($\pm$1.100e-02)} & 8 \\
$f_{6}$ & \textbf{\makecell{2.667e-01 \\ ($\pm$7.212e-02)}} & 1 & \makecell{5.179e-01 \\ ($\pm$1.255e-02)} & 3 & \makecell{7.166e-01 \\ ($\pm$3.708e-02)} & 5 & \makecell{4.807e-01 \\ ($\pm$1.274e-01)} & 2 & \makecell{9.681e-01 \\ ($\pm$7.910e-02)} & 8 & \makecell{9.274e-01 \\ ($\pm$4.337e-02)} & 7 & \makecell{8.168e-01 \\ ($\pm$4.617e-03)} & 6 & \makecell{6.944e-01 \\ ($\pm$2.639e-02)} & 4 & \makecell{9.771e-01 \\ ($\pm$1.505e-02)} & 9 \\
$f_{7}$ & \textbf{\makecell{3.874e-03 \\ ($\pm$6.870e-04)}} & 1 & \makecell{4.001e-03 \\ ($\pm$7.460e-04)} & 2 & \makecell{8.427e-02 \\ ($\pm$3.778e-02)} & 4 & \makecell{8.907e-03 \\ ($\pm$1.812e-02)} & 3 & \makecell{1.098e+00 \\ ($\pm$3.987e-01)} & 9 & \makecell{6.442e-01 \\ ($\pm$4.161e-01)} & 7 & \makecell{3.420e-01 \\ ($\pm$7.018e-03)} & 5 & \makecell{4.512e-01 \\ ($\pm$4.541e-03)} & 6 & \makecell{1.086e+00 \\ ($\pm$1.927e-01)} & 8 \\
$f_{8}$ & \textbf{\makecell{3.367e-01 \\ ($\pm$2.348e-02)}} & 1 & \makecell{8.715e-01 \\ ($\pm$5.092e-02)} & 5 & \makecell{9.1354e-01 \\ ($\pm$1.579e-02)} & 6 & \makecell{3.659e-01 \\ ($\pm$1.398e-01)} & 2 & \makecell{1.003e+00 \\ ($\pm$7.569e-03)} & 8 & \makecell{1.013e+00 \\ ($\pm$5.447e-02)} & 9 & \makecell{5.862e-01 \\ ($\pm$8.866e-03)} & 4 & \makecell{5.536e-01 \\ ($\pm$2.587e-02)} & 3 & \makecell{9.975e-01 \\ ($\pm$1.890e-02)} & 7 \\
$f_{9}$ & \textbf{\makecell{2.537e-02 \\ ($\pm$2.005e-03)}} & 1 & \makecell{5.137e-02 \\ ($\pm$1.062e-02)} & 2 & \makecell{4.501e-01 \\ ($\pm$7.413e-02)} & 4 & \makecell{1.066e-01 \\ ($\pm$1.178e-01)} & 3 & \makecell{9.306e-01 \\ ($\pm$1.456e-01)} & 8 & \makecell{8.636e-01 \\ ($\pm$8.021e-02)} & 7 & \makecell{7.126e-01 \\ ($\pm$3.002e-03)} & 6 & \makecell{4.778e-01 \\ ($\pm$1.119e-02)} & 5 & \makecell{9.913e-01 \\ ($\pm$3.306e-02)} & 9 \\
$f_{10}$ & \textbf{\makecell{1.976e-01 \\ ($\pm$3.647e-02)}} & 1 & \makecell{5.653e-01 \\ ($\pm$5.259e-02)} & 4 & \makecell{8.981e-01 \\ ($\pm$3.143e-02)} & 6 & \makecell{3.3089e-01 \\ ($\pm$1.677e-01)} & 2 & \makecell{9.989e-01 \\ ($\pm$1.288e-02)} & 9 & \makecell{9.756e-01 \\ ($\pm$3.716e-02)} & 7 & \makecell{5.990e-01 \\ ($\pm$7.860e-03)} & 5 & \makecell{5.445e-01 \\ ($\pm$2.943e-02)} & 3 & \makecell{9.862e-01 \\ ($\pm$2.242e-02)} & 8 \\
$f_{11}$ & \textbf{\makecell{1.519e-01 \\ ($\pm$2.166e-02)}} & 1 & \makecell{3.964e-01 \\ ($\pm$4.593e-02)} & 3 & \makecell{5.097e-01 \\ ($\pm$1.203e-01)} & 5 & \makecell{1.970e-01 \\ ($\pm$7.259e-02)} & 2 & \makecell{1.033e+00 \\ ($\pm$1.906e-01)} & 9 & \makecell{9.117e-01 \\ ($\pm$1.137e-01)} & 7 & \makecell{5.612e-01 \\ ($\pm$3.153e-03)} & 6 & \makecell{4.915e-01 \\ ($\pm$1.228e-02)} & 4 & \makecell{1.005e+00 \\ ($\pm$4.577e-02)} & 8 \\
$f_{12}$ & \textbf{\makecell{9.932e-03 \\ ($\pm$6.366e-04)}} & 1 & \makecell{2.868e-02 \\ ($\pm$9.360e-03)} & 3 & \makecell{2.879e-01 \\ ($\pm$5.871e-02)} & 4 & \makecell{2.485e-02 \\ ($\pm$3.037e-02)} & 2 & \makecell{1.063e+00 \\ ($\pm$2.957e-01)} & 9 & \makecell{8.441e-01 \\ ($\pm$1.329e-01)} & 7 & \makecell{5.733e-01 \\ ($\pm$4.341e-03)} & 6 & \makecell{4.559e-01 \\ ($\pm$1.932e-03)} & 5 & \makecell{9.745e-01 \\ ($\pm$5.487e-02)} & 8 \\
$f_{13}$ & \textbf{\makecell{1.049e-01 \\ ($\pm$9.260e-03)}} & 1 & \makecell{1.7164e-01 \\ ($\pm$1.487e-02)} & 2 & \makecell{4.8383e-01 \\ ($\pm$7.9395e-02)} & 4 & \makecell{3.137e-01 \\ ($\pm$1.479e-01)} & 3 & \makecell{1.001e+00 \\ ($\pm$2.541e-01)} & 8 & \makecell{9.594e-01 \\ ($\pm$2.338e-01)} & 7 & \makecell{6.662e-01 \\ ($\pm$5.119e-03)} & 6 & \makecell{5.596e-01 \\ ($\pm$1.274e-02)} & 5 & \makecell{1.0156e+00 \\ ($\pm$3.386e-02)} & 9 \\
$f_{14}$ & \textbf{\makecell{3.467e-01 \\ ($\pm$4.029e-02)}} & 1 & \makecell{5.102e-01 \\ ($\pm$3.6687e-02)} & 2 & \makecell{9.260e-01 \\ ($\pm$5.280e-02)} & 6 & \makecell{6.529e-01 \\ ($\pm$2.383e-01)} & 4 & \makecell{9.967e-01 \\ ($\pm$1.256e-02)} & 9 & \makecell{9.875e-01 \\ ($\pm$1.634e-02)} & 7 & \makecell{7.879e-01 \\ ($\pm$7.140e-03)} & 5 & \makecell{5.686e-01 \\ ($\pm$5.574e-02)} & 3 & \makecell{9.943e-01 \\ ($\pm$4.790e-03)} & 8 \\
$f_{15}$ & \textbf{\makecell{1.858e-02 \\ ($\pm$1.865e-03)}} & 1 & \makecell{3.8045e-02 \\ ($\pm$8.402e-03)} & 2 & \makecell{2.932e-01 \\ ($\pm$1.075e-01)} & 4 & \makecell{4.592e-02 \\ ($\pm$6.621e-02)} & 3 & \makecell{8.080e-01 \\ ($\pm$3.355e-01)} & 7 & \makecell{8.680e-01 \\ ($\pm$1.952e-01)} & 8 & \makecell{4.667e-01 \\ ($\pm$8.117e-03)} & 6 & \makecell{4.437e-01 \\ ($\pm$3.554e-03)} & 5 & \makecell{9.609e-01 \\ ($\pm$7.286e-02)} & 9 \\
$f_{16}$ & \textbf{\makecell{3.395e-02 \\ ($\pm$2.048e-03)}} & 1 & \makecell{7.089e-02 \\ ($\pm$1.301e-02)} & 3 & \makecell{3.765e-01 \\ ($\pm$4.461e-02)} & 4 & \makecell{6.115e-02 \\ ($\pm$3.722e-02)} & 2 & \makecell{1.126e+00 \\ ($\pm$3.399e-01)} & 9 & \makecell{9.115e-01 \\ ($\pm$2.915e-01)} & 7 & \makecell{4.315e-01 \\ ($\pm$8.638e-03)} & 5 & \makecell{4.688e-01 \\ ($\pm$3.911e-03)} & 6 & \makecell{9.796e-01 \\ ($\pm$6.266e-02)} & 8 \\
$f_{17}$ & \textbf{\makecell{1.074e-01 \\ ($\pm$1.976e-02)}} & 1 & \makecell{1.561e-01 \\ ($\pm$1.215e-02)} & 2 & \makecell{5.205e-01 \\ ($\pm$1.133e-01)} & 5 & \makecell{2.837e-01 \\ ($\pm$1.382e-01)} & 3 & \makecell{1.016e+00 \\ ($\pm$2.113e-01)} & 9 & \makecell{9.282e-01 \\ ($\pm$1.371e-01)} & 7 & \makecell{6.732e-01 \\ ($\pm$6.446e-03)} & 6 & \makecell{4.838e-01 \\ ($\pm$2.307e-03)} & 4 & \makecell{9.962e-01 \\ ($\pm$2.465e-02)} & 8 \\
$f_{18}$ & \textbf{\makecell{1.044e-02 \\ ($\pm$1.963e-03)}} & 1 & \makecell{1.062e-02 \\ ($\pm$1.989e-03)} & 2 & \makecell{8.199e-02 \\ ($\pm$5.328e-02)} & 4 & \makecell{1.191e-02 \\ ($\pm$1.388e-02)} & 3 & \makecell{1.212e+00 \\ ($\pm$5.077e-01)} & 9 & \makecell{1.007e+00 \\ ($\pm$6.887e-01)} & 8 & \makecell{3.406e-01 \\ ($\pm$1.511e-02)} & 5 & \makecell{4.496e-01 \\ ($\pm$6.134e-03)} & 6 & \makecell{9.946e-01 \\ ($\pm$1.919e-01)} & 7 \\
$f_{19}$ & \textbf{\makecell{3.679e-02 \\ ($\pm$2.354e-03)}} & 1 & \makecell{4.410e-02 \\ ($\pm$2.112e-03)} & 2 & \makecell{2.334e-01 \\ ($\pm$9.778e-02)} & 4 & \makecell{4.639e-02 \\ ($\pm$1.501e-02)} & 3 & \makecell{1.346e+00 \\ ($\pm$4.673e-01)} & 9 & \makecell{9.383e-01 \\ ($\pm$3.202e-01)} & 7 & \makecell{4.732e-01 \\ ($\pm$7.545e-03)} & 6 & \makecell{4.503e-01 \\ ($\pm$4.796e-03)} & 5 & \makecell{9.453e-01 \\ ($\pm$7.680e-02)} & 8 \\
$f_{20}$ & \makecell{8.398e-01 \\ ($\pm$2.960e-02)} & 3 & \makecell{1.083e+00 \\ ($\pm$1.905e-02)} & 7 & \makecell{1.012e+00 \\ ($\pm$2.183e-01)} & 6 & \makecell{1.331e+00 \\ ($\pm$2.046e-01)} & 9 & \makecell{1.112e+00 \\ ($\pm$1.961e-01)} & 8 & \makecell{9.420e-01 \\ ($\pm$1.718e-01)} & 4 & \makecell{6.981e-01 \\ ($\pm$5.944e-03)} & 2 & \textbf{\makecell{4.765e-01 \\ ($\pm$3.638e-03)}} & 1 & \makecell{9.996e-01 \\ ($\pm$3.267e-02)} & 5 \\
$f_{21}$ & \makecell{7.040e-01 \\ ($\pm$9.098e-02)} & 3 & \makecell{1.364e+00 \\ ($\pm$2.093e-01)} & 8 & \makecell{1.202e+00 \\ ($\pm$9.325e-01)} & 7 & \makecell{2.158e+00 \\ ($\pm$7.968e-01)} & 9 & \makecell{9.083e-01 \\ ($\pm$4.507e-01)} & 6 & \makecell{7.193e-01 \\ ($\pm$1.741e-01)} & 4 & \makecell{4.875e-01 \\ ($\pm$1.261e-02)} & 2 & \textbf{\makecell{4.644e-01 \\ ($\pm$4.439e-03)}} & 1 & \makecell{8.450e-01 \\ ($\pm$9.333e-02)} & 5 \\
$f_{22}$ & \textbf{\makecell{1.510e-01 \\ ($\pm$1.464e-02)}} & 1 & \makecell{2.461e-01 \\ ($\pm$4.918e-02)} & 2 & \makecell{5.732e-01 \\ ($\pm$5.330e-01)} & 5 & \makecell{6.557e-01 \\ ($\pm$4.232e-01)} & 6 & \makecell{1.218e+00 \\ ($\pm$2.861e-01)} & 9 & \makecell{7.922e-01 \\ ($\pm$2.258e-01)} & 7 & \makecell{4.764e-01 \\ ($\pm$1.002e-02)} & 3 & \makecell{4.877e-01 \\ ($\pm$7.833e-03)} & 4 & \makecell{9.235e-01 \\ ($\pm$1.072e-01)} & 8 \\
$f_{23}$ & \makecell{6.710e-01 \\ ($\pm$8.559e-02)} & 2 & \makecell{8.138e-01 \\ ($\pm$5.816e-02)} & 4 & \makecell{8.647e-01 \\ ($\pm$1.863e-01)} & 5 & \makecell{1.203e+00 \\ ($\pm$2.712e-01)} & 9 & \makecell{9.253e-01 \\ ($\pm$7.969e-02)} & 6 & \makecell{9.422e-01 \\ ($\pm$1.104e-01)} & 7 & \makecell{7.416e-01 \\ ($\pm$6.220e-03)} & 3 & \textbf{\makecell{5.120e-01 \\ ($\pm$6.008e-03)}} & 1 & \makecell{9.840e-01 \\ ($\pm$1.687e-02)} & 8 \\
$f_{24}$ & \textbf{\makecell{4.208e-02 \\ ($\pm$3.578e-03)}} & 1 & \makecell{4.976e-02 \\ ($\pm$4.706e-03)} & 2 & \makecell{3.044e-01 \\ ($\pm$7.671e-02)} & 4 & \makecell{1.205e-01 \\ ($\pm$2.249e-01)} & 3 & \makecell{8.704e-01 \\ ($\pm$3.136e-01)} & 7 & \makecell{8.725e-01 \\ ($\pm$1.594e-01)} & 8 & \makecell{5.949e-01 \\ ($\pm$1.245e-02)} & 6 & \makecell{4.478e-01 \\ ($\pm$4.802e-03)} & 5 & \makecell{9.443e-01 \\ ($\pm$3.974e-02)} & 9 \\
$f_{25}$ & \textbf{\makecell{4.761e-01 \\ ($\pm$4.270e-03)}} & 1 & \makecell{5.228e-01 \\ ($\pm$1.212e-02)} & 2 & \makecell{7.801e-01 \\ ($\pm$8.267e-02)} & 5 & \makecell{7.829e-01 \\ ($\pm$1.751e-01)} & 6 & \makecell{1.074e+00 \\ ($\pm$1.468e-01)} & 9 & \makecell{9.507e-01 \\ ($\pm$7.340e-02)} & 7 & \makecell{7.639e-01 \\ ($\pm$6.876e-03)} & 4 & \makecell{5.5681e-01 \\ ($\pm$3.762e-02)} & 3 & \makecell{9.924e-01 \\ ($\pm$2.6151e-02)} & 8 \\
$f_{26}$ & \textbf{\makecell{8.457e-02 \\ ($\pm$2.309e-03)}} & 1 & \makecell{1.247e-01 \\ ($\pm$6.232e-03)} & 2 & \makecell{5.769e-01 \\ ($\pm$8.750e-02)} & 5 & \makecell{3.5784e-01 \\ ($\pm$2.156e-01)} & 3 & \makecell{1.070e+00 \\ ($\pm$1.034e-01)} & 9 & \makecell{9.564e-01 \\ ($\pm$1.231e-01)} & 7 & \makecell{7.449e-01 \\ ($\pm$7.159e-03)} & 6 & \makecell{5.134e-01 \\ ($\pm$2.629e-02)} & 4 & \makecell{1.002e+00 \\ ($\pm$2.409e-02)} & 8 \\
$f_{27}$ & \textbf{\makecell{2.105e-02 \\ ($\pm$1.565e-03)}} & 1 & \makecell{4.306e-02 \\ ($\pm$8.079e-03)} & 3 & \makecell{2.923e-01 \\ ($\pm$5.900e-02)} & 4 & \makecell{2.9602e-02 \\ ($\pm$2.527e-02)} & 2 & \makecell{9.909e-01 \\ ($\pm$4.818e-01)} & 9 & \makecell{8.536e-01 \\ ($\pm$2.447e-01)} & 7 & \makecell{4.469e-01 \\ ($\pm$1.127e-02)} & 5 & \makecell{4.675e-01 \\ ($\pm$4.484e-03)} & 6 & \makecell{9.740e-01 \\ ($\pm$8.670e-02)} & 8 \\
$f_{28}$ & \textbf{\makecell{2.587e-01 \\ ($\pm$1.918e-01)}} & 1 & \makecell{4.070e-01 \\ ($\pm$1.711e-01)} & 3 & \makecell{1.111e+00 \\ ($\pm$9.875e-01)} & 8 & \makecell{6.8929e+00 \\ ($\pm$9.001e+00)} & 9 & \makecell{8.864e-01 \\ ($\pm$5.261e-01)} & 6 & \makecell{5.730e-01 \\ ($\pm$3.018e-01)} & 5 & \makecell{3.956e-01 \\ ($\pm$1.691e-02)} & 2 & \makecell{5.343e-01 \\ ($\pm$2.068e-02)} & 4 & \makecell{1.047e+00 \\ ($\pm$1.295e-01)} & 7 \\
$f_{29}$ & \textbf{\makecell{8.247e-02 \\ ($\pm$3.261e-02)}} & 1 & \makecell{2.027e-01 \\ ($\pm$6.516e-02)} & 2 & \makecell{3.869e-01 \\ ($\pm$2.315e-01)} & 4 & \makecell{9.670e-01 \\ ($\pm$7.828e-01)} & 8 & \makecell{1.450e+00 \\ ($\pm$1.488e+00)} & 9 & \makecell{8.228e-01 \\ ($\pm$5.330e-01)} & 6 & \makecell{2.846e-01 \\ ($\pm$1.410e-02)} & 3 & \makecell{4.086e-01 \\ ($\pm$1.100e-02)} & 5 & \makecell{9.128e-01 \\ ($\pm$2.151e-01)} & 7 \\
$f_{30}$ & \textbf{\makecell{1.550e-01 \\ ($\pm$4.946e-02)}} & 1 & \makecell{4.503e-01 \\ ($\pm$1.825e-01)} & 4 & \makecell{8.709e-01 \\ ($\pm$7.084e-01)} & 7 & \makecell{1.643e+00 \\ ($\pm$1.535e+00)} & 9 & \makecell{1.066e+00 \\ ($\pm$5.893e-01)} & 8 & \makecell{6.208e-01 \\ ($\pm$4.472e-01)} & 5 & \makecell{3.211e-01 \\ ($\pm$1.223e-02)} & 2 & \makecell{4.434e-01 \\ ($\pm$1.118e-02)} & 3 & \makecell{8.194e-01 \\ ($\pm$1.122e-01)} & 6 \\
$f_{31}$ & \textbf{\makecell{4.195e-01 \\ ($\pm$1.646e-02)}} & 1 & \makecell{5.097e-01 \\ ($\pm$1.885e-02)} & 2 & \makecell{7.354e-01 \\ ($\pm$7.600e-02)} & 5 & \makecell{6.307e-01 \\ ($\pm$1.275e-01)} & 3 & \makecell{9.029e-01 \\ ($\pm$5.651e-02)} & 7 & \makecell{9.083e-01 \\ ($\pm$6.827e-02)} & 8 & \makecell{7.724e-01 \\ ($\pm$6.649e-03)} & 6 & \makecell{6.314e-01 \\ ($\pm$2.295e-02)} & 4 & \makecell{9.753e-01 \\ ($\pm$1.479e-02)} & 9 \\
$f_{32}$ & \textbf{\makecell{4.606e-02 \\ ($\pm$2.856e-03)}} & 1 & \makecell{7.069e-02 \\ ($\pm$3.246e-03)} & 2 & \makecell{3.982e-01 \\ ($\pm$7.023e-02)} & 4 & \makecell{1.703e-01 \\ ($\pm$8.434e-02)} & 3 & \makecell{1.023e+00 \\ ($\pm$1.880e-01)} & 9 & \makecell{9.262e-01 \\ ($\pm$1.439e-01)} & 7 & \makecell{6.723e-01 \\ ($\pm$1.213e-02)} & 6 & \makecell{4.737e-01 \\ ($\pm$1.424e-02)} & 5 & \makecell{9.577e-01 \\ ($\pm$3.192e-02)} & 8 \\ \hline
\textbf{Avg Rank} & \multicolumn{2}{c}{\textbf{1.15625}} & \multicolumn{2}{c}{2.84375} & \multicolumn{2}{c}{4.9375} & \multicolumn{2}{c}{4.0625} & \multicolumn{2}{c}{8.25} & \multicolumn{2}{c}{6.875} & \multicolumn{2}{c}{4.875} & \multicolumn{2}{c}{4.125} & \multicolumn{2}{c}{7.875} \\ \hline
\end{tabular}
}
\end{table*}

The numerical results and performance rankings of Meta-DO and its competitors across the 32 benchmark instances are summarized in \Cref{tab:full_res_on_benchmark}. The following observations can be drawn:
1) \textbf{Overall Superiority}: The proposed Meta-DO consistently dominates the SOTA baselines across the majority of the heterogeneous test instances. Our approach achieves the best mean performance in 29 of 32 cases and maintains the highest average rank of 1.15625. This demonstrates the effectiveness of the meta-policy in automating the detect-then-act pipeline, allowing the optimizer to respond to environmental changes without manual parameter re-tuning.
2) \textbf{Comparison with Specialized Baselines}: Compared to representative multi-population frameworks (mCMAES, mDE) and clust\-ering-based methods (APCPSO, DPCPSO), Meta-DO demonstrates superior adaptability. While these baselines rely on rigid, human-crafted reactive logic and fixed thresholds to trigger environmental responses, Meta-DO leverages population-level features to autonomously derive optimal control parameters, thus bridging the gap between static heuristics and dynamic requirements.
3) \textbf{Robustness in Volatile and Coupled Landscapes}: On highly challenging instances involving landscape switching ($f_{15}\text{-}$ $f_{24}$) and hybrid transitions ($f_{25}\text{-}f_{32}$), where the search space is both unpredictable and structurally complex, the performance gap between Meta-DO and baselines becomes even more pronounced. This underscores the robustness of our reinforcement learning-assisted approach in accurately tracking moving optima under extreme uncertainty and coupled topological complexities.
\subsection{Generalization Analysis} \label{sec:okayplan}
To further validate the practical efficacy and real-time performance of the proposed framework beyond synthetic benchmark functions, we evaluate our algorithm on a complex dynamic path planning task. This experiment is based on the OkayPlan benchmark \cite{XinOkayPlan}, which simulates the navigation of an Unmanned Surface Vehicle (USV) in scenarios with multiple moving obstacles.
The rationale for selecting this practical task is three-fold: 
1) \textbf{High Real-time Requirement}: Unlike static optimization, dynamic path planning requires the optimizer to generate collision-free trajectories at a high frequency to respond to the kinematic changes of obstacles. This serves as a demanding test for our Meta-DO's computational efficiency. 
2) \textbf{Dynamic Obstacle Avoidance}: The problem is formulated as an Obstacle Kinematics Augmented Optimization Problem (OKAOP). The optimizer must continuously track the safe regions in a rapidly shifting search space, testing our framework's adaptability.
3) \textbf{Generalization to Engineering Tasks}: By testing the model (pre-trained on synthetic datasets) on this domain-specific problem, we can assess the cross-domain generalization capability of the learned strategy.

To quantitatively evaluate the performance of our framework in dynamic navigation, we employ three key performance indicators:
1) \textbf{Success Rate (SR)}: The percentage of episodes where the robot successfully reaches the target area without colliding with any dynamic obstacles. This metric reflects the overall reliability and safety of the planner.
2) \textbf{Distance to Target ($D_{\text{target}}$)}: The Euclidean distance between the robot's final position and the center of the target at the end of an episode. A smaller $D_{\text{target}}$ indicates higher terminal precision and a more effective guidance strategy.
3) \textbf{Time Steps ($T_{\text{step}}$)}: The total number of discrete control cycles (frames) taken by the robot until the episode terminates. This value represents the path efficiency and real-time responsiveness of the optimizer in dynamic scenarios.
\subsubsection{Experimental Settings and Scenario Configurations}
The experimental scenarios are categorized into six cases with varying levels of environmental uncertainty and task complexity, as summarized below:

$\bullet$ Case 1-3 (Consistent Obstacles): These cases involve obstacles with predictable, consistent kinematics. Complexity scales with 4, 6, and 10 segments, respectively.

$\bullet$ Case 4-6 (Random Obstacles): To evaluate the optimizer's responsiveness to stochastic environments, these cases utilize obstacles with randomized motion patterns. Complexity similarly scales across 4, 6, and 10 segments.

To simulate strict real-time constraints, we impose the following operational limits on all tested algorithms: 
1) \textbf{Temporal Constraint}: Each navigation episode is limited to at most 500 frames. If the USV fails to reach the target within this threshold, the episode is marked as a failure.
2) \textbf{Computational Budget}: Within each discrete frame, the optimizer is permitted a maximum of 1,000 FEs to determine the optimal next-state velocity and heading. This fixed budget setting ensures a fair comparison of the algorithms' efficiency in generating real-time control commands. 

The frame-by-frame update mechanism requires the optimizer to provide a valid, collision-free solution within milliseconds, directly challenging the meta-policy's ability to balance exploration and exploitation under extreme time pressure.

\subsubsection{Experiments result}
\begin{figure*}[t]
    \centering
    \includegraphics[width=0.85\textwidth]{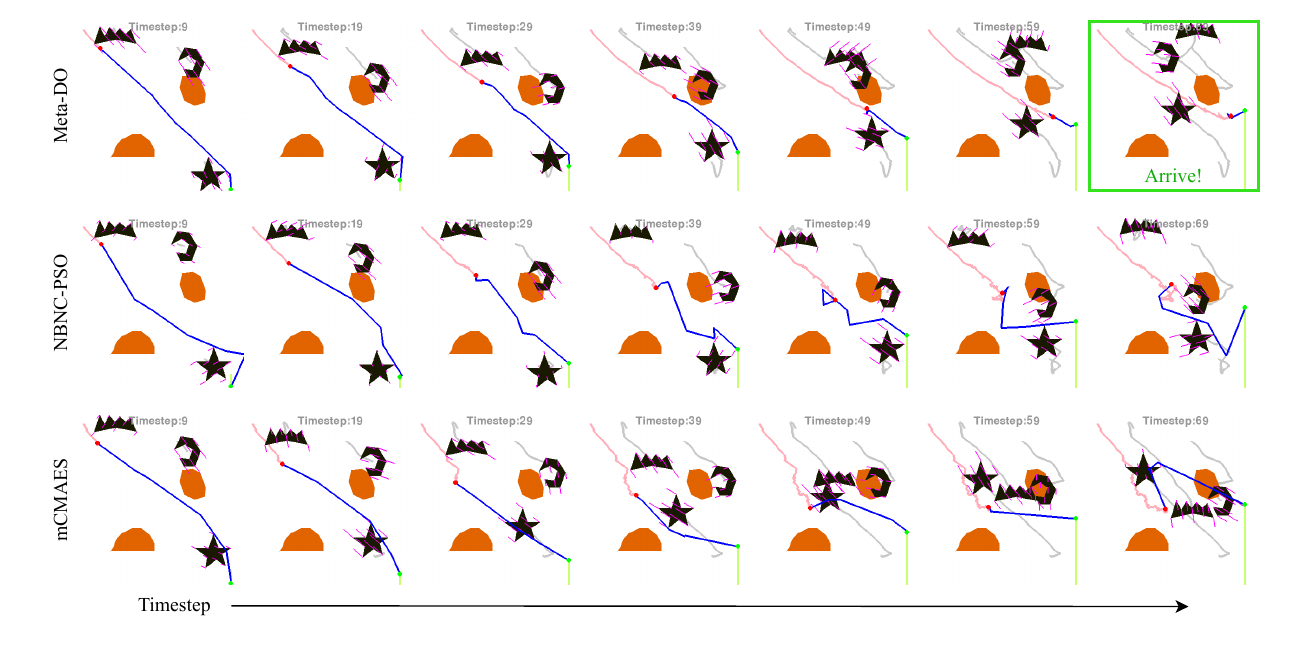}
    \vspace{-5mm}
    \caption{Snapshots of the simulation results on case 6}
    \label{Fig.okayplan}
\end{figure*}

Some path trajectories are shown in \Cref{Fig.okayplan}.
Without any domain-specific retraining, the meta-policy (pre-trained solely on synthetic functions) successfully guides the USV through complex moving obstacles. Due to space constraints, \Cref{tab:res_on_okayplan} compares the base optimizer (NBNC-PSO) and the most competitive SOTA baseline mCMAES. The comprehensive results involving all eight baseline algorithms are provided in Appendix \ref{appx:B} %B 
for a more exhaustive evaluation. As shown in \Cref{tab:res_on_okayplan}, our approach maintains significantly higher Success Rates (SR) compared to both the backbone NBNC-PSO and competitor mCMAES.

Under strict computational budgets and real-time constraints, Meta-DO achieves superior terminal precision ($D_{\text{target}}$) and path efficiency ($T_{\text{step}}$). This demonstrates our framework can generalize its learned policy to high-frequency engineering applications.

\begin{table}[htbp]
\centering
\caption{Comparative Results on Real-time Path Planning (Representative Baselines)}
\label{tab:res_on_okayplan}
\resizebox{\columnwidth}{!}{
\begin{tabular}{c | ccc | ccc | ccc } \hline
\text{Algorithm} & \multicolumn{3}{c|}{\textbf{Meta-DO}} & \multicolumn{3}{c|}{NBNC-PSO} & \multicolumn{3}{c}{mCMAES} \\ \hline
Metrics & SR & $D_{\text{target}}$ & $T_{\text{step}}$ & SR & $D_{\text{target}}$ & $T_{\text{step}}$ & SR & $D_{\text{target}}$ & $T_{\text{step}}$ \\ \hline
case 1 & \textbf{0.3} & \textbf{78.40} & 11.18 & 0.0 & 168.66 & 10.60 & 0.1 & 287.66 & 10.31 \\
case 2 & \textbf{0.5} & \textbf{122.15} & \textbf{11.73} & 0.0 & 220.56 & 17.26 & 0.2 & 167.41 & 16.18 \\
case 3 & \textbf{0.7} & \textbf{67.65} & 11.89 & 0.0 & 179.03 & \textbf{10.95} & 0.2 & 118.98 & 23.21 \\
case 4 & \textbf{0.7} & \textbf{95.40} & 7.47 & 0.1 & 128.03 & 10.43 & 0.0 & 322.48 & 6.15 \\
case 5 & \textbf{0.9} & \textbf{34.86} & \textbf{14.82} & 0.0 & 209.27 & 20.62 & 0.2 & 173.00 & 16.67 \\
case 6 & \textbf{0.9} & \textbf{21.85} & \textbf{20.51} & 0.0 & 186.81 & 20.87 & 0.0 & 138.23 & 24.09 \\ \hline
\end{tabular}
}
\end{table}
\begin{figure}[t]
    \centering
    \includegraphics[width=0.85\columnwidth]{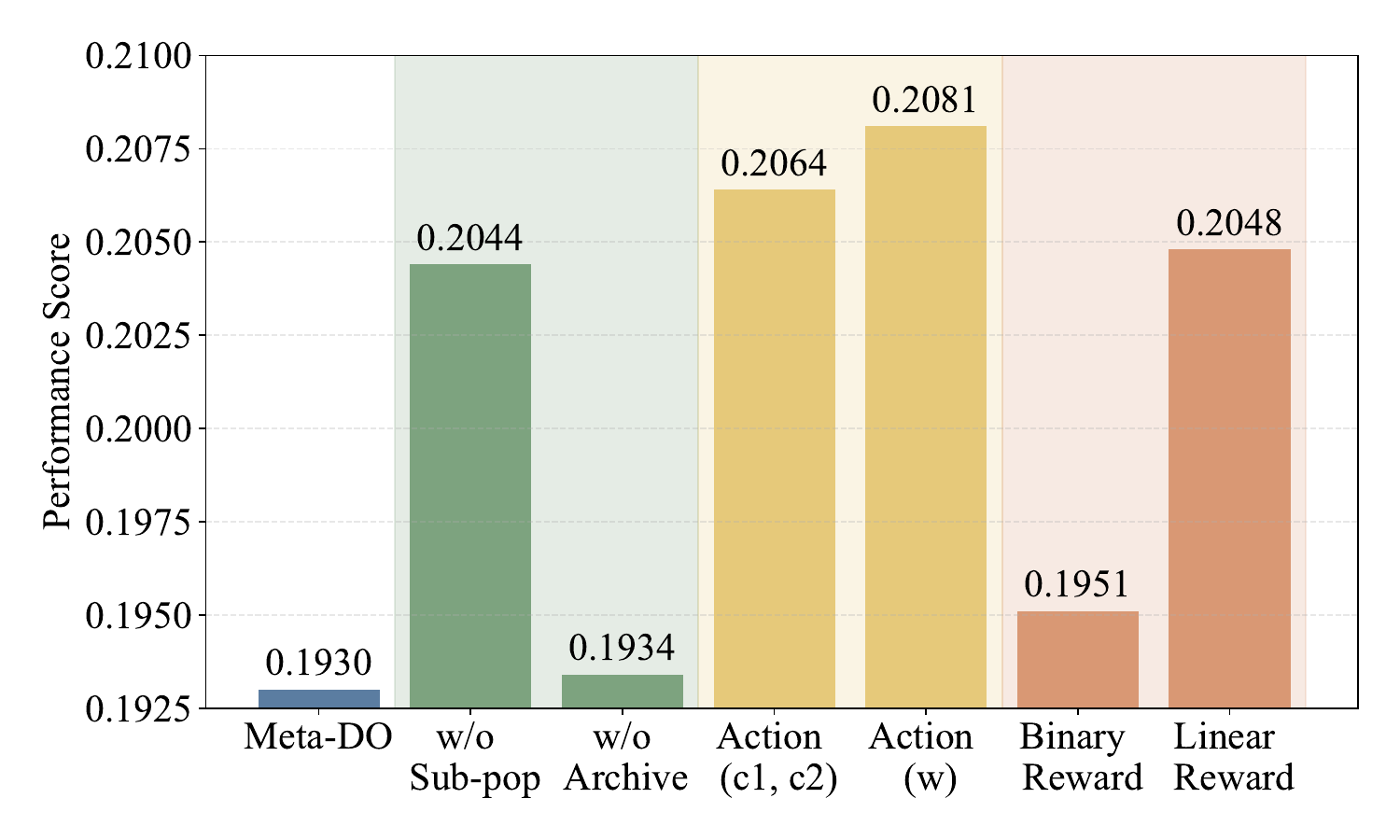}
    \vspace{-3mm}
    \caption{Ablation study results of Meta-DO}
    \vspace{-5mm}
    \label{Fig.barchart}
\end{figure}

\subsection{Ablation study}
To rigorously evaluate the contribution of each design element in Meta-DO, we compare the full framework against six degraded variants. These variants are designed to test the necessity of our state representation, action space, and reward design.
\begin{itemize}
\item[1)] State Representation Variants:
a) \textbf{w/o Sub-pop}: This variant removes the local context features from state vector ($\text{fea}_3$ and $\text{fea}_8$).
b) \textbf{w/o Archive}: Removes the archive from state ($\text{fea}_1$), testing the importance of long-term memory in tracking shifting optima.
\item[2)] Action Space Variants:
a) \textbf{Action ($c_1, c_2$)}: The RL agent only controls the acceleration coefficients. To ensure a fair baseline, the inertia weight $w$ is set to follow a standard linear decay: $w=0.9-0.5\cdot(\text{generation}/T_{max})$.
b) \textbf{Action ($w$)}: The agent only regulates $w$, while the acceleration coefficients are fixed at the commonly used value of $c_1 = c_2 = 2.05$.
\item[3)] Reward Mechanism Variants:
a) \textbf{Binary Reward}: Replaces the continuous log-scale reward with a sparse signal ($r=1$ for any fitness improvement, $r=0$ otherwise).
b) \textbf{Linear Reward}: Uses the absolute fitness improvement as the reward signal without $\log$ transformation, assessing the framework's sensitivity to fitness scales.
\end{itemize}

The results are present in \Cref{Fig.barchart}:
1) \textbf{Importance of Sub-popu\-lation Features}: Removing the local context features ($fea_3$ and $fea_8$) in the "w/o Sub-pop" variant leads to a noticeable performance decline. This confirms that niching-based state information is indispensable for the agent to perceive local convergence status and manage multiple species effectively.
2) \textbf{Role of Long-term Memory}: Lacking historical best solution status ($fea_1$), the "w/o Archive" variant struggles to track optima across rapid transitions. This highlights the necessity of environmental variation perception for maintaining stability in non-stationary landscapes.
3) \textbf{Reward Scaling \& Action Space}: The comparison with Binary and Linear Reward variants demonstrates that our log-scale reward mechanism is essential for providing a stable and scale-invariant learning signal. Furthermore, the joint control of all three hyper-parameters ($w, c_1, c_2$) proves superior to partial control, as it allows the agent to fully regulate the exploration-exploitation balance.

\section{Conclusion}
In this paper, we present Meta-DO, a RL-assisted framework designed to enable automated detection of environmental changes and self-adaptation in DOPs. By streamlining the traditional detect-then-act pipeline into an end-to-end MDP, the current optimization state can be mapped directly to strategy adjustments. This eliminates the reliance on hand-crafted change detection mechanisms and manual response heuristics, enabling the low-level optimizer to robustly track shifting optima in non-stationary landscapes.
Experimental results on 32 heterogeneous instances verify that Meta-DO significantly outperforms SOTA baselines and demonstrates favorable generalization capabilities across different problem classes. Moreover, the practical efficacy and cross-domain robustness of the learned policy are further validated through a real-time USV path planning task. 
However, there are still some limitations in this study. The current low-level searching behavior is constrained by the predefined PSO paradigm, and the population-level features may be unstable in high-dimensional scenarios. Future work involves investigating individual-level operator adaptation and learning-to-optimize paradigms to further enhance the performance and flexibility of dynamic-aware EC algorithms.

% \begin{acks}
% xiexie
% \end{acks}

\bibliographystyle{ACM-Reference-Format}
\bibliography{sample-base}
% \iffalse
\newpage
\appendix
\section{Experimental Details} \label{appx:A}
Meta-DO is implemented using PyTorch and trained on a 13th Gen Intel Core i7-13700H CPU. The meta-policy is optimized for 20 epochs using the Adam optimizer with a learning rate of $1 \times 10^{-5}$ for both the actor and critic networks. Both training and testing batch sizes are set to 8.
\section{Real-time Path Planning Details}\label{appx:B}
As introduced in \Cref{sec:okayplan}, the OkayPlan benchmark evaluates the cross-domain generalization of our framework. While synthetic benchmarks focus on mathematical landscapes, this USV navigation task introduces physical constraints and high-frequency real-time requirements. The purpose of this appendix is to provide a comprehensive comparison against all eight SOTA baselines, including advanced clustering methods (DPCPSO, APCPSO) and specialized dynamic optimizers (PSPSO, DynDE), to further validate the robustness of the learned Meta-DO policy.

\begin{table*}[htbp]
\centering
\caption{Comparative Results on Real-time Path Planning}
\label{tab:full_res_on_okayplan}
\resizebox{\textwidth}{!}{
\begin{tabular}{c | ccc | ccc | ccc | ccc | ccc | ccc | ccc | ccc | ccc} \hline
\text{Algorithm} & \multicolumn{3}{c|}{\textbf{Meta-DO}} & \multicolumn{3}{c|}{NBNC-PSO} & \multicolumn{3}{c|}{PSPSO} & \multicolumn{3}{c|}{ACFPSO} & \multicolumn{3}{c|}{mCMAES} & \multicolumn{3}{c|}{mDE} & \multicolumn{3}{c|}{APCPSO} & \multicolumn{3}{c|}{DPCPSO} & \multicolumn{3}{c}{DynDE} \\ \hline
Metrics & SR & $D_{\text{target}}$ & $T_{\text{step}}$ & SR & $D_{\text{target}}$ & $T_{\text{step}}$ & SR & $D_{\text{target}}$ & $T_{\text{step}}$ & SR & $D_{\text{target}}$ & $T_{\text{step}}$ & SR & $D_{\text{target}}$ & $T_{\text{step}}$ & SR & $D_{\text{target}}$ & $T_{\text{step}}$ & SR & $D_{\text{target}}$ & $T_{\text{step}}$& SR & $D_{\text{target}}$ & $T_{\text{step}}$ & SR & $D_{\text{target}}$ & $T_{\text{step}}$ \\ \hline
case 1 & \textbf{0.3} & \textbf{78.40} & 11.18 & 0 & 168.66 & 10.60 & 0 & 287.51 & \textbf{4.28} & 0 & 334.71 & 4.89 & 0.1 & 287.66 & 10.31 & 0 & 284.60 & 15.05 & 0 & 364.00 & 6.59 & 0.1 & 284.44 & 40.00 & 0 & 320.56 & 99.95 \\
case 2 & \textbf{0.5} & \textbf{122.15} & 11.73 & 0 & 220.56 & 17.26 & 0.1 & 222.64 & \textbf{9.70} & 0 & 292.90 & 17.74 & 0.2 & 167.41 & 16.18 & 0 & 331.46 & 21.18 & 0 & 397.12 & 11.63 & 0.3 & 198.12 & 73.99 & 0 & 350.56 & 235.40 \\
case 3 & \textbf{0.7} & \textbf{67.65} & 11.89 & 0 & 179.03 & 10.95 & 0.1 & 191.30 & 17.16 & 0 & 245.82 & 12.36 & 0.2 & 118.98 & 23.21 & 0 & 223.70 & 16.96 & 0 & 386.71 & \textbf{5.17} & 0.1 & 206.98 & 36.49 & 0 & 307.86 & 214.12 \\
case 4 & \textbf{0.7} & \textbf{95.40} & 7.47 & 0.1 & 128.03 & 10.43 & 0.1 & 270.03 & 8.74 & 0 & 332.24 & 12.20 & 0 & 322.48 & 6.15 & 0 & 312.71 & 16.31 & 0.1 & 359.07 & \textbf{3.53} & 0 & 303.76 & 64.23 & 0 & 304.58 & 68.60 \\
case 5 & \textbf{0.9} & \textbf{34.86} & 14.82 & 0 & 209.27 & 20.62 & 0.1 & 230.07 & 13.96 & 0 & 270.69 & 47.27 & 0.2 & 173.00 & 16.67 & 0 & 301.15 & 27.36 & 0 & 390.68 & \textbf{6.62} & 0.3 & 206.79 & 40.35 & 0 & 346.89 & 270.39 \\
case 6 & \textbf{0.9} & \textbf{21.85} & 20.51 & 0 & 186.81 & 20.87 & 0 & 181.25 & 19.02 & 0 & 244.57 & 22.03 & 0 & 138.23 & 24.09 & 0 & 223.92 & 22.18 & 0 & 362.60 & \textbf{7.83} & 0 & 250.36 & 49.92 & 0 & 340.52 & 223.10 \\ \hline
\end{tabular}
}
\end{table*}
\Cref{tab:full_res_on_okayplan} presents the complete performance metrics across all six test cases. Several critical observations can be made from the extended baseline data:
\begin{itemize}
\item[$\bullet$] Failure of Heuristic Clustering: Recent clustering-based algorithms like DPCPSO and APCPSO, which perform well on synthetic functions, show nearly zero Success Rates (SR) in the most challenging case 4-6. This confirms that fixed clustering thresholds are insufficient for the non-stationary topology found in obstacle-augmented spaces
\item[$\bullet$] Safety-Efficiency Trade-off: In case 1, PSPSO achieves a very low $T_{\text{step}}$ (4.28), but its SR is 0, indicating a failed navigation. Meta-DO is the only framework that maintains a stable balance between path safety and efficiency.
\item[$\bullet$] Robustness in Stochasticity: In cases 4-6 where obstacle motion is randomized, the gap between Meta-DO and all SOTA baselines widens significantly. Meta-DO maintains an SR of 0.7-0.9 while most baselines fail, demonstrating that the agent effectively learned a generalized environment-response mechanism.
\end{itemize}

The extended results confirm that Meta-DO's superior performance holds against a wide array of DOAs. The ability to maintain a high success rate (SR) under strict computational budgets underscores the practical potential of our proposed framework in real-world engineering.
% \fi
\end{document}